%% file: emnlp2022.tex
\pdfoutput=1

\documentclass[11pt]{article}

\usepackage{EMNLP2022}

\usepackage{times}
\usepackage{latexsym}

\usepackage[T1]{fontenc}

\usepackage[utf8]{inputenc}

\usepackage{microtype}

\usepackage{inconsolata}

\usepackage{amsmath}
\usepackage{graphicx}
\usepackage{amsfonts}
\usepackage{xcolor}
\usepackage{multirow}
\usepackage{booktabs}
\usepackage{natbib}
\usepackage{mathtools}
\usepackage{caption}
\usepackage{kotex}
\usepackage{mathtools, nccmath}
\usepackage{algorithm2e}
\usepackage[bottom]{footmisc}

\DeclarePairedDelimiter{\nint}\lfloor\rceil

\include{macros}

\title{Quadapter: Adapter for GPT-2 Quantization}

\author{{\bf Minseop Park, Jaeseong You, Markus Nagel, Simyung Chang} \\ Qualcomm AI Research\thanks{\phantom{c}Qualcomm AI Research is an initiative of Qualcomm Technologies, Inc.} \\  \texttt{\{minspark, jaeseong, markusn, simychan\}@qti.qualcomm.com}}

\begin{document}
\maketitle
\begin{abstract}

Transformer language models such as GPT-2 are difficult to quantize because of outliers in activations leading to a large quantization error. To adapt to the error, one must use quantization-aware training, which entails a fine-tuning process based on the dataset and the training pipeline identical to those for the original model. Pretrained language models, however, often do not grant access to their datasets and training pipelines, forcing us to rely on arbitrary ones for fine-tuning. In that case, it is observed that quantization-aware training overfits the model to the fine-tuning data. 
For quantization without overfitting, we introduce a quantization adapter (Quadapter), a small set of parameters that are learned to make activations quantization-friendly by scaling them channel-wise. It keeps the model parameters unchanged.
By applying our method to the challenging task of quantizing GPT-2, 
we demonstrate that it effectively prevents the overfitting and improves the quantization performance.

\end{abstract}

\section{Introduction}




Quantizing a transformer model is not a simple matter due to numerous channel-dependent outliers in activations \cite{DBLP:conf/emnlp/BondarenkoNB21}. They lead to a large quantization error \cite{DBLP:conf/icml/ZhaoHDSZ19}, and we observe that the problem is worse in the decoder-only transformers like GPT-2. One solution to the difficulty is quantization-aware training (QAT), an approach that fine-tunes the model parameters in response to the numerical error arising from quantization. Post-training quantization (PTQ) -- a counterpart of QAT that performs quantization without modifying model parameters -- is not powerful enough to cope with the outliers.


\begin{figure}[!ht]
    \centering
    \includegraphics[width=1.\linewidth]{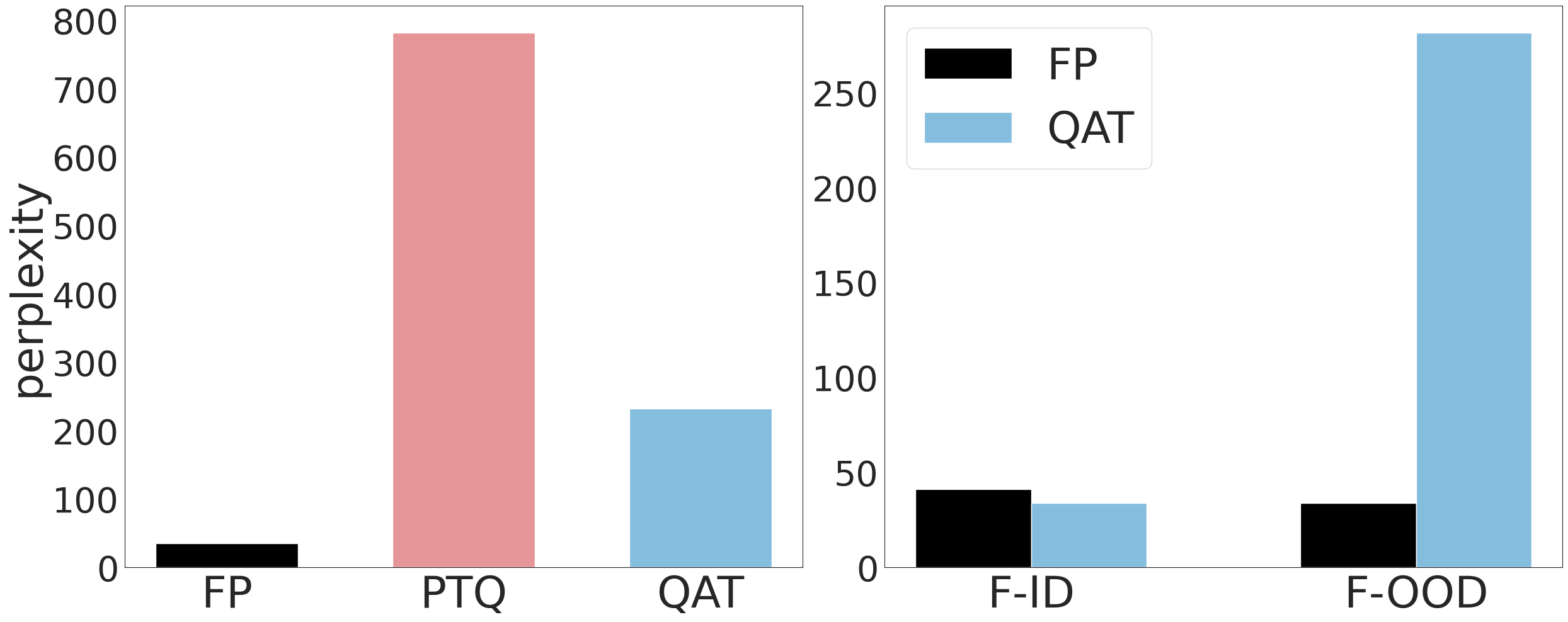}
    \vskip -0.1in
    \caption{
    Average perplexity (PPL) of 
    the full-precision (FP) model and the models quantized with PTQ and QAT on 
    5 datasets (left). 
    We use the PTB dataset as the fine-tuning data (F-ID) for QAT. 
    The FP model and the QAT model are evaluated on the F-ID and the other 4 datasets (F-OOD) (right).
    }
    \label{motivation}
\end{figure}


While QAT is effective, it requires the dataset and the training pipeline, and the problem is that they are often inaccessible when dealing with the original pretrained model without any downstream task. One then cannot but use arbitrary fine-tuning data for QAT. 

However, the fine-tuning returns worse accuracies for distributions unseen during training (out-of-distribution with regard to fine-tuning; F-OOD) despite improving for the training distribution (in-distribution with regard to fine-tuning; F-ID) \cite{DBLP:journals/corr/abs-2202-10054}.
This is consistent with our observation that QAT overfits the model to the fine-tuning data as in Figure~\ref{motivation}.
The resulting quantized model therefore has its generality impaired. This violates the premise of a general-purpose language model, which must operate well across various texts of the target language. 


\begin{figure*}
    \centering
    \includegraphics[width=6in]{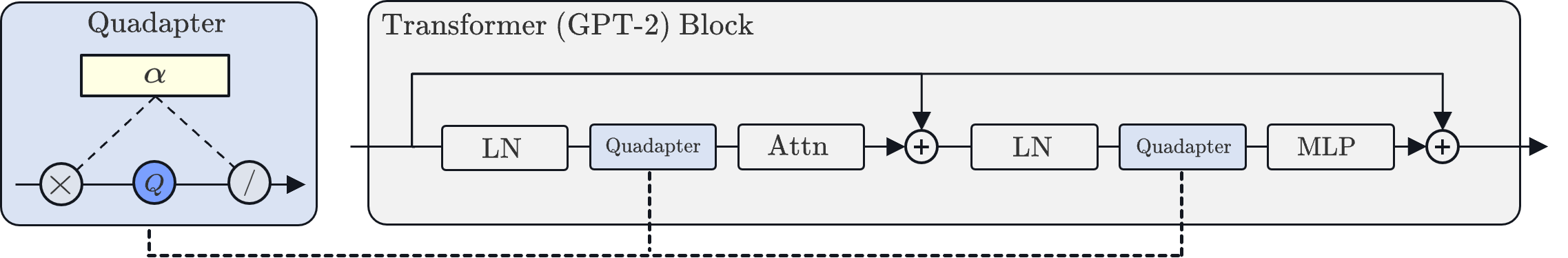}
    \vskip -0.05in
    \caption{
    Quadapter performs a linear scaling and its inversion before and after $\tQ$, the quantizer for the target activation (left).
    In the transformer block of GPT-2, Quadapters can be installed in two different locations (right).
    }
    \label{fig:quadapter}
\end{figure*}

Our hypothesis is that QAT incurs the overfitting because it changes all the parameters of the model. This difficulty is much like the research topic of continual learning, where it is important that a model should not forget its past capability when transferring to a new task \cite{DBLP:journals/corr/abs-2111-00667}. Adapter is a strategy to adapt to a new distribution by training only a small number of parameters. It is a popular method to lessen the catastrophic forgetting. We borrow this concept to propose Quadapter, a lightweight module to adapt to the quantization error on behalf of the intact original model.



The contribution of this work is that we successfully quantize GPT-2, overcoming the large inter-channel variance and the QAT overfitting issue with Quadapter. To the best of our knowledge, this is the first work to quantize both weights and activations of GPT-2 
without the complete training pipeline.
%

\section{Related Works}

\noindent{\textbf{Adapters}} 
Extensive researches have been conducted on how to steer a large pretrained model with few adapter parameters.
The concept of adapter has proven its usefulness in language models for
transfer learning \cite{pmlr-v97-houlsby19a}, 
multi-task learning \cite{DBLP:journals/corr/abs-1902-02671},
and domain adaptation \cite{DBLP:journals/corr/abs-2111-00667}. 
Several works apply adapters to the visual domain as well
\cite{DBLP:conf/eccv/LiH16, DBLP:conf/aaai/PerezSVDC18}. 

\noindent{\textbf{Transformer Quantization}}
In comparison to GPT-2,
BERT is easier to quantize. It can be quantized
with PTQ under a limited performance drop \cite{DBLP:conf/aaai/ShenDYMYGMK20}.
QAT on BERT for a given downstream task
recovers full-precision (FP) performance even with ultra-low precision \cite{DBLP:conf/nips/ZafrirBIW19, DBLP:conf/emnlp/BondarenkoNB21}, or with 
integer-only operations for non-linear layers \cite{DBLP:conf/icml/KimGYMK21}.
On the other hand,
quantization studies on autoregressive transformers are relatively limited in their scope, using weight-only quantization
\cite{DBLP:journals/corr/abs-2009-07453} or requiring full-fledged training \cite{DBLP:journals/corr/abs-1910-10485,DBLP:conf/acl/taoacl2022}. Please note that these works focus on quantizing GPT-2 that is finetuned on a downstream task whereas ours quantizes the original pretrained GPT-2.



\noindent{\textbf{Quantization techniques}}
Directly relevant to our work are cross-layer-equalization (CLE) \cite{9008784} and adaptive rounding (AdaRound) \cite{DBLP:conf/icml/NagelABLB20}. Similarly to CLE, Qudapter rescales associated model weights to lessen the quantization burden. AdaRound and our proposed method are alike in training foldable helper parameters to minimize the block-wise quantization error. 
In addition, learned step size (LSQ) \cite{DBLP:conf/iclr/EsserMBAM20} and its extension (LSQ+) \cite{9151058} train the quantization-related parameters during QAT, to which Quadapter bears similarity. 

\begin{table*}
    \centering
    \resizebox{1.0\textwidth}{!}{
    \begin{tabular}{llccccc|ccccc} \toprule
        & & \multicolumn{5}{c}{GPT-2} & \multicolumn{5}{c}{DistilGPT-2}\\
        Data & Method & Wikitext2 & PTB & LAMBADA & CBT\_CN & CBT\_NE  & Wikitext2 & PTB & LAMBADA & CBT\_CN & CBT\_NE \\
        \hline\hline
        - & FP32 & 29.27 & 41.31 & 48.39 & 27.29 & 30.53 & 44.36 & 59.73 & 74.94 & 42.54 & 47.09 \\
        \hline
        \multirow{4}{*}{Calib. data}
         & PTQ & 915.58 & 751.23 & 827.06 & 655.31 & 759.83 & 87.52 & 114.42 & 205.35 & 93.16 & 104.94 \\
         & AdaRound & 507.07 & 478.29 & 685.98 & 319.74 & 309.11 & 84.94 & 104.94 & 164.98 & 107.89 & 92.59 \\
         & CLE & 40.28  & 59.33 & 74.61  & 38.92  & 43.69 & 69.81 & 86.66 & 144.06 & 68.78 & 76.80 \\
         & Quadapter BC & \textbf{34.53} & \textbf{50.65} & \textbf{63.51} & \textbf{32.47} & \textbf{36.46} & \textbf{52.79} & \textbf{70.43} & \textbf{102.75} & \textbf{51.97} & \textbf{57.81} \\
        \hline\hline
        \multirow{3}{*}{Wikitext2}
         & QAT & \underline{32.51}   & 100.75   & 125.40   & 54.94    & 63.94 & \underline{35.04} & 109.40 & 129.19 & 67.03 & 76.55   \\
         & Quadapter BC+QAT & \underline{\textbf{21.61}} & 57.06    & 63.65    & 33.80    & 38.40 & \underline{\textbf{28.50}} & 80.52 & 86.57 & 50.64 & 57.05   \\
         & Quadapter (ours) & \underline{29.34}   & \textbf{47.30} & \textbf{57.28} & \textbf{30.37} & \textbf{34.05} & \underline{43.05} & \textbf{66.28} & \textbf{85.42} & \textbf{47.66} & \textbf{52.49}  \\
        \hline
        \multirow{3}{*}{PTB}
         & QAT & 331.61    & \underline{33.94}     & 330.10    & 212.12    & 252.03  & 347.25 & \underline{37.44} & 308.22 & 214.14 & 257.44 \\
         & Quadapter BC+QAT & 79.74     & \underline{\textbf{24.10}} & 106.32    & 59.90     & 69.79  & 121.62 & \underline{\textbf{29.65}} & 146.48 & 91.73 & 106.31  \\
         & Quadapter (ours) & \textbf{33.69} & \underline{39.46}     & \textbf{55.68} & \textbf{31.45} & \textbf{35.16} & \textbf{50.73} & \underline{56.63} & \textbf{87.02} & \textbf{49.43} & \textbf{54.35}  \\
        \bottomrule
    \end{tabular}
    }
    \vskip -0.1in
    \caption{
    Performance evaluation of the quantized GPT-2 and DistilGPT-2 on various datasets. The metric is PPL (lower is better).
    In the case of Quadapter BC+QAT, QAT initiates after the block-wise calibration of Quadapter. For Quadapter (ours), both the training phases are completed. \underline{Underline} indicates the results on F-ID 
    }
    \label{tab:main}
\end{table*}

\section{Methods}

Quadapter is simply a set of learnable parameters. 
On the other hand, the Quadapter block represents the actual working mechanism of Quadapter, involving two consecutive layers of linear relations, their quantizers, and their associated Quadapter instance. The effectiveness of Quadapter comes from the interaction amongst the involved components, and from the two-phase training procedure.

\subsection{Quadapter Design}
Quadatper linearly scales the input channels and reverts after quantization. This ensures the identity relation if not for quantizers, making it possible to keep the model parameters intact (Figure~\ref{fig:quadapter} left).

The scaling and the inverse-scaling of an activation are, in practice, folded to the weight and the bias of the preceding layer and to the weight of the following layer. For example, given a forward pass of two linear layers:
\begin{align}
\by & = \bW_2(\bW_1\bx + \bb_1)+\bb_2,
\label{eq:fpforward}
\end{align}
the Quadapter block output $\hat\by$ is as follows:
\begin{align}
    \hat\by = \tQ_{\btheta_2}(\bW_2\bA^{-1}) \tQ_{\btheta_a}(\tQ_{\btheta_1}( \bA\bW_1)\bx  \nonumber \\
    + \bA\bb_1) + \bb_2 \label{eq:main_1eq} \\
    = \tQ_{\btheta_2}(\bW_2')\tQ_{\btheta_a}(\tQ_{\btheta_1}(\bW_1')\bx + \bb_1') + \bb_2.
\label{eq:main_2eq}
\end{align}

Here, $\bA = \rm{diag}(\ba)$ is a diagonal matrix with $\bA_{ii}=\ba_i$, 
where $\alpha \in \bbR^d$ is the learnable Quadapter parameter with the intermediate activation dimension $d$.
$\tQ_{\btheta_1}$ and $\tQ_{\btheta_2}$ are the weight quantizers, 
and $\tQ_{\btheta_a}$ is the activation quantizer. 
Each quantizer $\tQ_\btheta$ quantizes its input values based on the quantization parameter 
$\btheta = (\theta_{{\rm min}}, \theta_{{\rm max}})$ \cite{DBLP:journals/corr/abs-1806-08342}. 
Quadapter $\ba$ is trained during training and fused at the inference time (Equation~\ref{eq:main_2eq}).





As in Equation~\ref{eq:main_1eq}, the forward scaling and the inverse scaling correspond across three nested quantizers that are strongly nonlinear operations. Therefore $\ba$ should be learned rather than set analytically as in \cite{9008784}; a single analytical solution is not sufficient to balance the quantization burden between the two layers.

\subsection{Quadapter Training}
The learning of Quadapter is comprised of two phases: the block-wise calibration and the end-to-end fine-tuning. 

\noindent{\textbf{Phase 1: Block-wise Calibration }}
Each of the Quadapter instances is initialized to $\vec{\mathbf{1}}$ and trained with the calibration data, independently per Quadpter block.
The local objective for each block is L2 loss:
\begin{align}
    \argmin_{\ba} ||\by - \hat\by||_2^2,
\end{align}
which \cite{DBLP:conf/icml/NagelABLB20} shows to be effectively complementary to the task loss.

$\hat\by$ is computed in the dynamic quantization mode \cite{DBLP:conf/nips/ZafrirBIW19}, where the statistics are obtained per batch.
Quadapter resulting from the calibration phase is a PTQ method that is independent of the fine-tuning process. We therefore denote such Quadapter by \textit{Quadapter BC}.


\noindent{\textbf{Phase 2: End-to-end Fine-tuning}}
The subsequent fine-tuning starts with more accommodating quantization parameters (i.e. the min/max statistics) since they have moved to moderate values from extreme outliers during the first phase. The fine-tuning therefore converges much more quickly.

In the second phase, the statistics for quantization are computed in the fashion of static quantization \cite{DBLP:conf/nips/ZafrirBIW19}, based on the same calibration data as in the first phase.
Quadapter is then trained to minimize the end-to-end task loss. During the course, the quantization parameters are jointly learned as in \cite{9151058} while the model parameters stay fixed. 
Algorithm \ref{algo1} details the full flow of the Quadatper training. 


\SetKwComment{Comment}{/* }{ */}
\RestyleAlgo{ruled}
\begin{algorithm}[hb!]
\caption{Quadapter training}\label{alg:two}
\SetKwData{Left}{left}\SetKwData{This}{this}\SetKwData{Up}{up}
\SetKwFunction{Union}{Union}\SetKwFunction{FindCompress}{FindCompress}
\SetKwInOut{Input}{input}\SetKwInOut{Output}{output}
\small
\Input{pretrained model $M$, Quadapter blocks, calibration data $D_1$, fine-tuning data $D_2$, learning rates $\eta_1$, $\eta_2$.}
\Output{
Learned Quadapter parameters $\{\ba_1, \ba_2, ...\}$
and quantization parameters $\btheta^* = \{\btheta^1, \btheta^2, ...\}$.
}
\Comment{Phase 1}
\ForEach{$i$-th Quadapter block}{
Initialize $\ba_i = \vec{\mathbf{1}}$ \\
From $M$ and $D_1$, gather block input $\bx_i$ and output $\by_i$ \\
\While{not converged}{
    $\hat\by_i \gets  \text{Eq. 2} $ \\
    $\ba_i \gets \ba_i - \eta_1 \nabla_{\ba_i} ||\hat\by_i  - \by_i||^2_2$ \\
}
}
\Comment{Phase 2}
Apply learned Quadapters to $M$ \\
Initialize $\btheta^*$ with $D_1$ to make quantized model $M_Q$ \\
\While{not converged}{
    \For{$\bx, \by \in D_2$}{
        compute $L_{\rm{task}}(M_Q(\bx), \by)$ \\
        \ForEach{$i$-th Quadapter block}{
            $\ba_i \gets \ba_i - \eta_2 \nabla_{\ba_i} L_{\rm{task}}$ \\
        }
        update $\boldsymbol\theta^*$ with LSQ+ \\
    }
}
\label{algo1}
\end{algorithm}

\section{Experiments}

\noindent{\textbf{Models}} We quantize GPT-2 \cite{radford2019language} and DistilGPT-2 \cite{sanh2019distilbert} based on their huggingface pretrained models\footnote{huggingface.co/gpt2, huggingface.co/distilgpt2}.
Our quantization configuration follows \cite{DBLP:journals/corr/abs-2201-08442}, doing uniform asymmetric 8-bit quantization for both activations and weights. All the weights and activations are quantized, except for biases, non-linear opererations, and additions \cite{DBLP:conf/nips/ZafrirBIW19, DBLP:conf/icml/KimGYMK21}. 
For every transformer block, the Quadapter instances are installed in between the first layer norm and the linear projection for key/query/value as well as between the second layer norm and the first feed-forward network (Figure \ref{fig:quadapter} right). One additional instance is applied to between the final layer norm and the logit projection.

\noindent{\textbf{Baseline methods}} Our implementation of LSQ+ follows the original proposition \cite{9151058}, except for updating the min/max parameters for stability of training \cite{DBLP:journals/corr/abs-2201-08442}. It is applied for all the QAT experiments. We use AI Model Efficiency Toolkit\footnote{https://github.com/quic/aimet} to obtain AdaRound performance. The CLE metrics are computed with an untrained Quadapter, initialized analytically as in \cite{9008784}.

\noindent{\textbf{Datasets}} We employ WikiText-2 \cite{merity2016pointer}, the English Penn Treebank (PTB) corpus \cite{DBLP:journals/coling/MarcusSM94}, the LAMBADA dataset \cite{DBLP:conf/acl/PapernoKLPBPBBF16}, and the named-entity subset (CBT\_NE) as well as the common-noun subset (CBT\_CN) of Children’s Book Test \cite{hill2016goldilocks}. We follow the datasets' default divisions as to training/validation/test splits.

\noindent{\textbf{Experiment design}} To test the overfitting resiliency, GPT-2 and DistilGPT-2 are quantized with various PTQ and QAT methods on one of the five datasets. The resulting quantized model is evaluated on its F-ID and on the other four datasets (F-OOD). In addition, we expose the models to varying amounts of fine-tuning data during quantization to compare the changing behaviors of QAT and Quadapter.

\begin{figure}[t]
    \includegraphics[width=\linewidth]{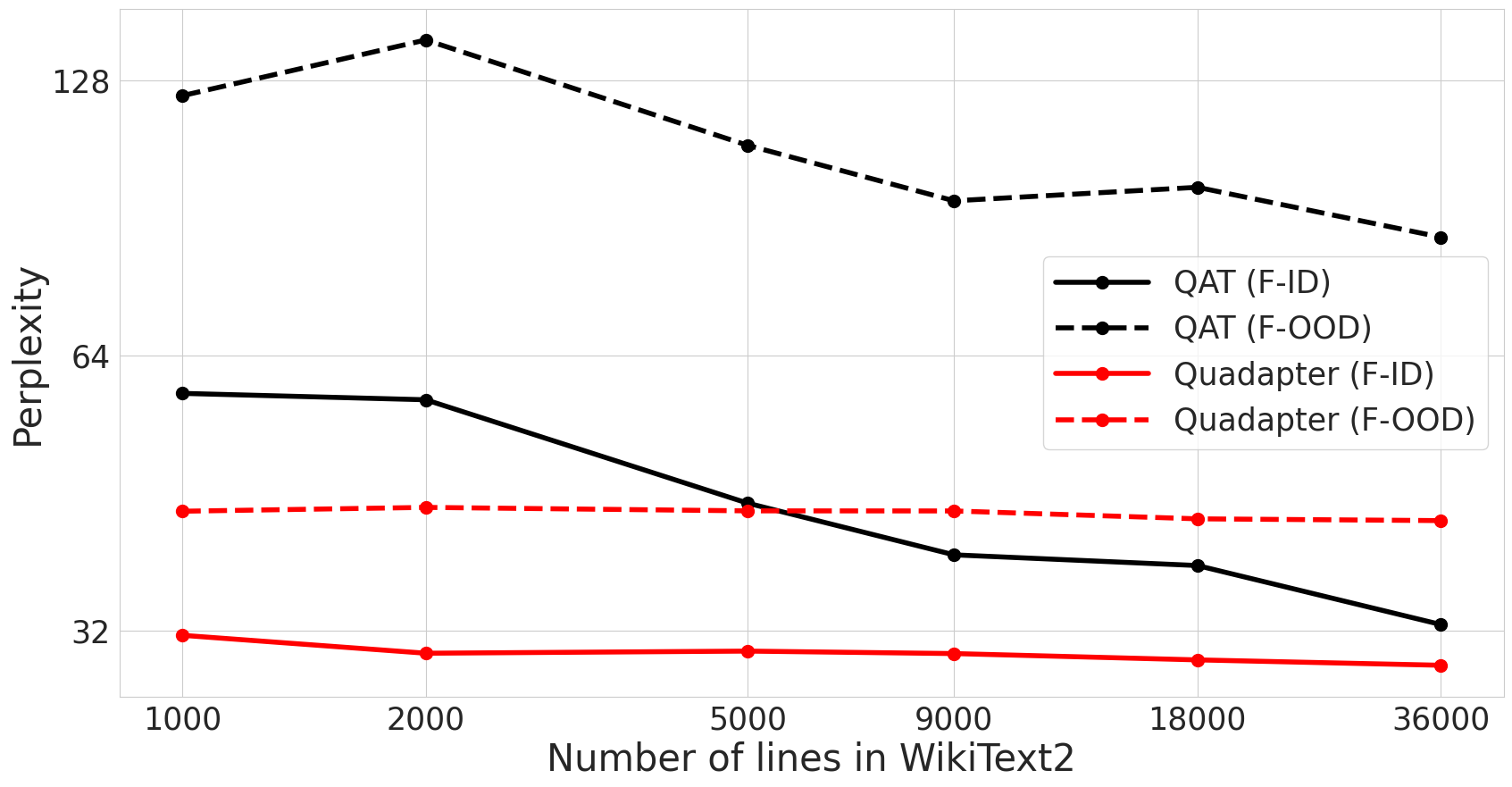}
    \vskip -0.05in
    \caption{
    GPT-2 quantization performance when fine-tuned on F-ID of varying sizes. Both axes are logarithmic.
    }
    \label{fig:num_line}
\end{figure}

\noindent{\textbf{Results}} In Table~\ref{tab:main}, Quadapter outperforms the baseline methods on the F-OOD in both GPT-2 and DistilGPT-2. This observation evinces the general capability of Quadapter to reduce overfitting across different models. The comparison between Quadapter (ours) and Quadapter BC+QAT is the ablation of the end-to-end finetuning, and the reusult proves its importance.

Noteworthy is that Quadapter is a powerful stand-alone PTQ technique. Even without QAT fine-tuning, the F-OOD metrics are better than those of the QAT baselines. In addition, the effectiveness of the calibration phase is shown by the comparison between CLE and Quadapter BC.


Another advantage of Quadapter is that it is a viable quantization option in data-scarce situations. 
As shown in Figure~\ref{fig:num_line}, Quadapter outperforms QAT throughout different amounts of fine-tuning data, and the gap is most evident when only a small amount of data is available. 

\begin{figure}[t]
    \centering
    \includegraphics[width=0.95\linewidth]{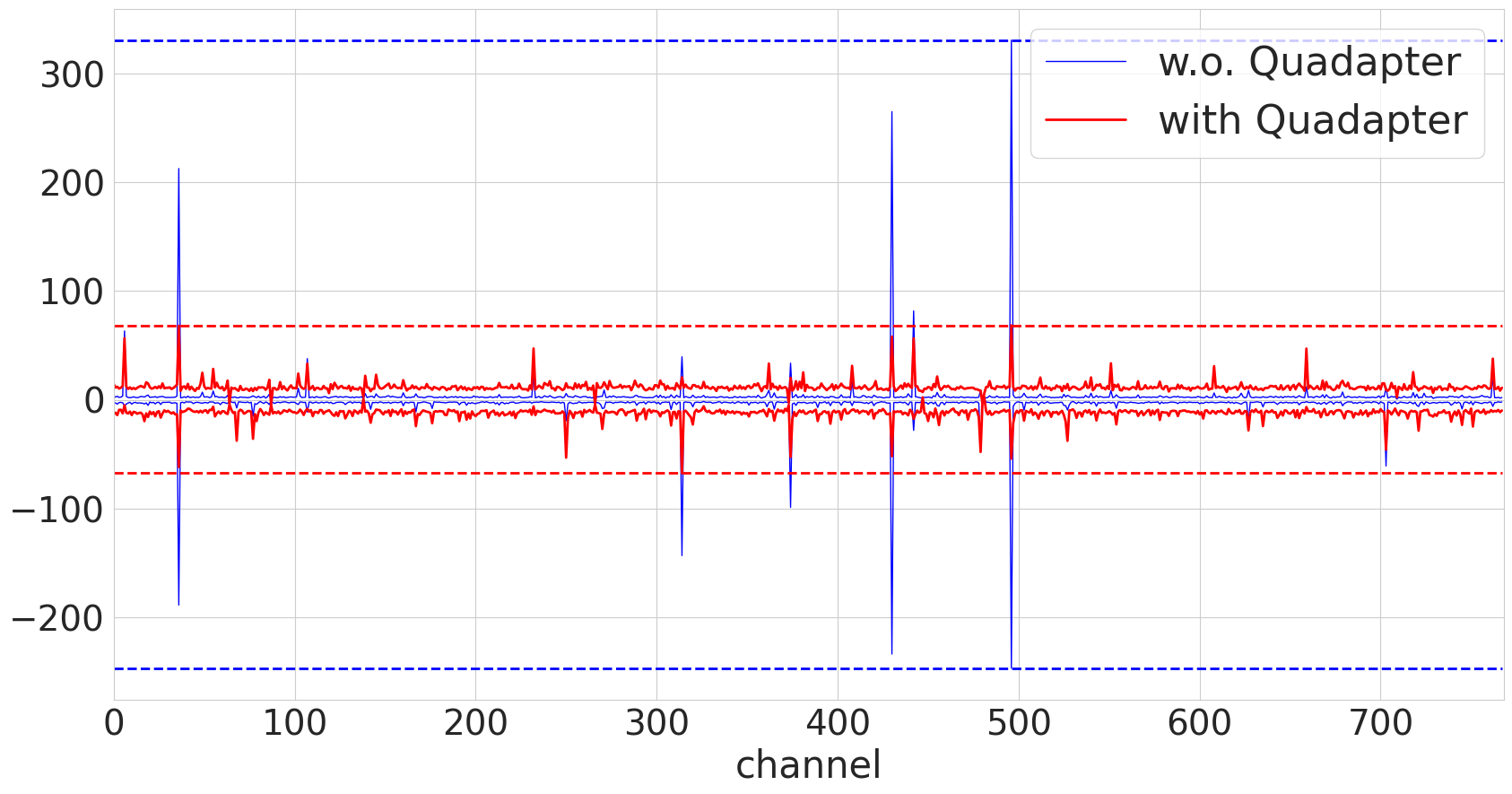}
    \vskip -0.1in
    \caption{Visualization of the per-channel (x-axis) min/max (y-axis) values of the final layer norm output activation in GPT-2. 
    The solid/dotted lines represent per-channel/total min and max.
    }
    \label{fig:stat}
\end{figure}

Aside from the convincing metrics reported above, we further explore if Quadapter does the intended job of transforming an activation into a more uniform distribution. Figure~\ref{fig:stat} describes the per-channel statistics before and after the Quadapter training. Values in most activation dimensions except for few have small magnitudes around 0, and such dimensions lose precision when quantized because of the large magnitudes of total min/max before applying Quadpater. The illustration verifies that the effect of Quadapter indeed aligns with our expectation, reducing the ranges of outlier-ridden channels while enlarging the ranges of the others.

\section{Limitations}
One limitation of Quadapter is that it requires two consecutive layers of linear relations. In other words, it can be a mediator only for convolution layers, linear layers, or normalization layers (when followed by a linear or convolution layer), but not if residual connections or nonlinear activation functions intervene. 

\section{Conclusions}
We identify two challenges in quantizing autoregressive transformer language models: the overfitting issue of QAT and the inter-channel variance in activations. Through experiments, we demonstrate that Quadapter not only mitigates the two problems but also serves as an effective PTQ technique.



\bibliographystyle{acl_natbib}
\bibliography{custom}

\clearpage

\appendix

\section*{Appendix}

\section{Additional Experiments}
\noindent{\textbf{F-ID Expansion}}
In Table~\ref{tab:main}, we limit the F-ID to one amongst the five available datasets. Here, we perform an additional experiment by expanding the F-ID to include four of them, limiting the F-OOD to the one remaining dataset. The results are in Table~\ref{tab:abl4food}. Comparing the metrics on WikiText2 when the QAT model is fine-tuned on PTB (Table~\ref{tab:main}) and when fine-tuned on all but WikiText2 (Table~\ref{tab:abl4food}), we can observe the improvement of Quadapter's F-OOD performance. On the other hand, QAT still suffers from overfitting despite the expanded fine-tuning data.

\noindent{\textbf{Ablation of LSQ+}}
In \cite{9151058}, LSQ+ is a composite method that includes initialization of weight quantizer, model parameter training, and quantization parameter ($\btheta$) training. However, in our work, we isolate the quantization parameter training and denote it by LSQ+. In Table~\ref{tab:main}, QAT is accompanied by LSQ+, thus training both the model parameters and the quantization parameters. We ablate LSQ+ in Table~\ref{tab:lsqabl}. The results show that LSQ+ tends to improve the quantization performance in general, 
particularly in conjunction with our proposed method.

\begin{figure}[htp]
    \centering
    \includegraphics[width=\linewidth]{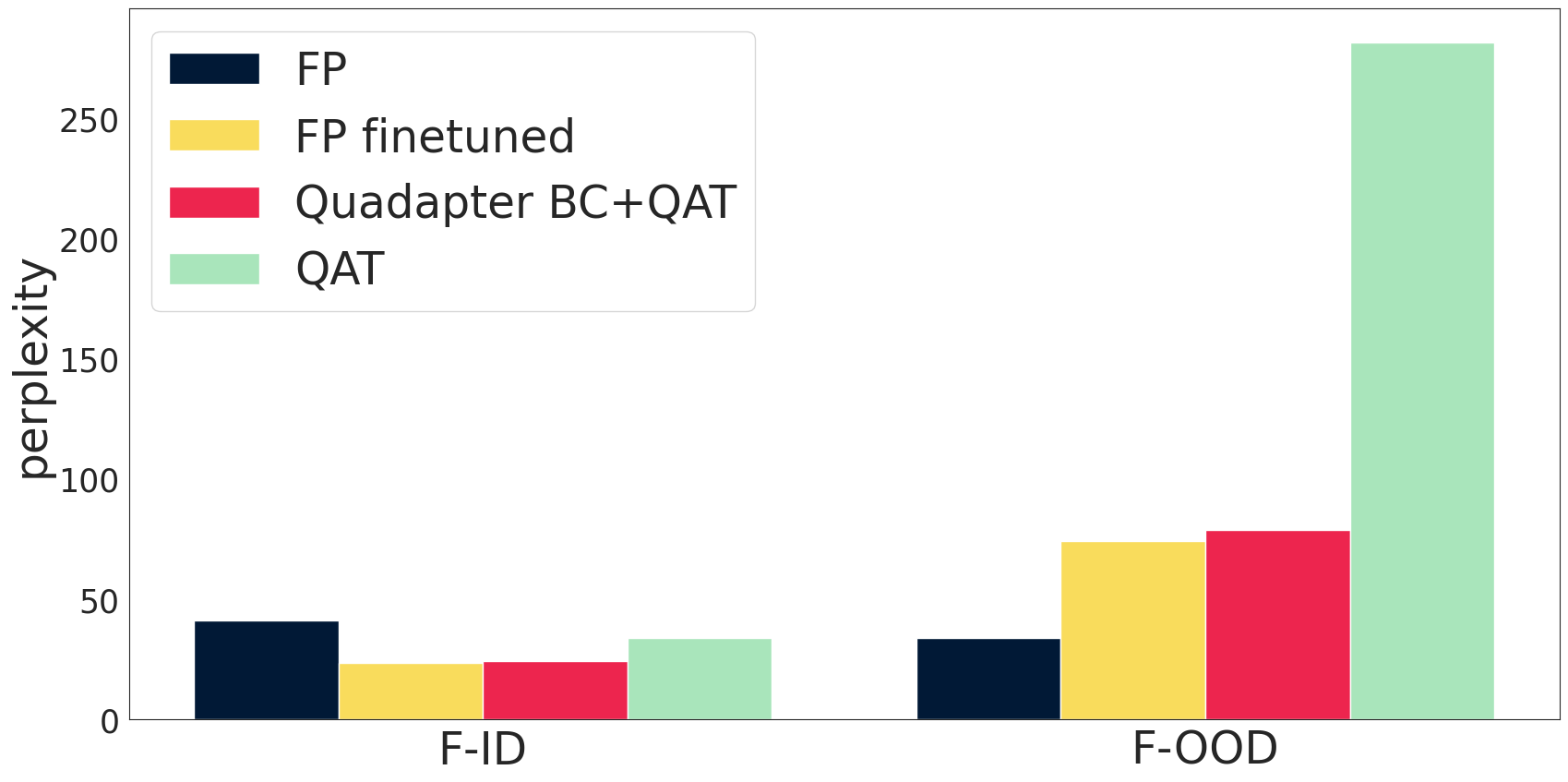}
    \caption{Comparison of the fine-tuned FP model (FP finetuned) with other methods. 
    PTB is the F-ID, and the other 4 datasets are the F-OOD.
    }
    \label{fig:fp_abl}
\end{figure}

\noindent{\textbf{Quadapter BC effectiveness on QAT}}
As discussed in the main text, QAT makes the model overfit to F-ID and perform poorly on F-OOD. 
However, when employed with Quadapter BC (i.e. Quadapter BC+QAT), the QAT training process is stabilized, and so the quantized model reaches near the upper bound of the fine-tuned FP model (Figure~\ref{fig:fp_abl}). This shows that Quadapter fosters QAT.

\begin{table*}
    \centering
    \resizebox{1.0\textwidth}{!}{
    \begin{tabular}{lccccc|cc} \toprule
         & Wikitext2 & PTB & LAMBADA & CBT\_CN & CBT\_NE & F-ID & F-OOD \\
         \hline\hline
         FP &  29.28 & 41.31 & 48.39 & 27.29 & 30.53 & 36.88 & 29.28 \\
         \hline
         QAT & 61.57 & 31.95 & 39.60 & 22.67 & 24.83 & 29.76 & 61.57 \\
         Quadapter BC+QAT & 41.13 & 25.37 & 34.05 & 19.79 & 21.56 & 25.19 & 41.13 \\
         Quadapter (ours) & 31.71 & 44.83 & 45.79 & 27.23 & 30.07 & 36.98 & 31.71 \\
         \bottomrule
    \end{tabular}
    }
    \caption{PPL measurements of GPT-2 for expanded F-ID. Wikitext2 is the F-OOD, and the other 4 datasets are the F-ID. The average PPL is reported in the columns, F-ID and F-OOD.}
    \label{tab:abl4food}
\end{table*}

\begin{table*}
    \centering
    \resizebox{1.0\textwidth}{!}{
    \begin{tabular}{llccccc} \toprule
    Data & Method & Wikitext2 & PTB & LAMBADA & CBT\_CN & CBT\_NE \\
    \hline\hline
    \multirow{3}{*}{Wikitext2} & QAT & 36.28 (3.77) & 101.81 (1.07) & 134.80 (9.40) & 58.77 (3.83) & 68.67 (4.73) \\
     & Quadapter BC+QAT & 21.63 (0.02) & 57.40 (0.34) & 61.72 (-1.93) & 33.48 (-0.32) & 38.02 (-0.38) \\
     & Qudapter (ours) & 32.69 (3.35) & 49.01 (1.71) & 59.59 (2.31) & 31.44 (1.07) & 35.27 (1.22) \\
    \hline
    \multirow{3}{*}{PTB} & QAT & 275.80 (-55.81) & 37.87 (3.93) & 284.61 (-45.49) & 165.75 (-46.37) & 194.71 (-57.32) \\
     & Quadapter BC+QAT & 80.00 (0.26) & 23.93 (-0.17) & 101.61 (-4.71) & 59.95 (0.05) & 69.78 (-0.01) \\
     & Quadapter (ours) & 33.79 (0.10) & 46.98 (7.52) & 59.46 (3.78) & 31.54 (0.09) & 35.37 (0.21) \\
    \bottomrule
    \end{tabular}
    }
    \caption {PPL measurements of GPT-2 trained without LSQ+.
    The differences from the counterparts with LSQ+ in Table~\ref{tab:main} are noted in the parenthesis (a positive value indicates LSQ+'s performance gain).}
    \label{tab:lsqabl}
\end{table*}

\section{Hyperparameter Details}
\begin{figure}[htp]
    \centering
    \includegraphics[width=\linewidth]{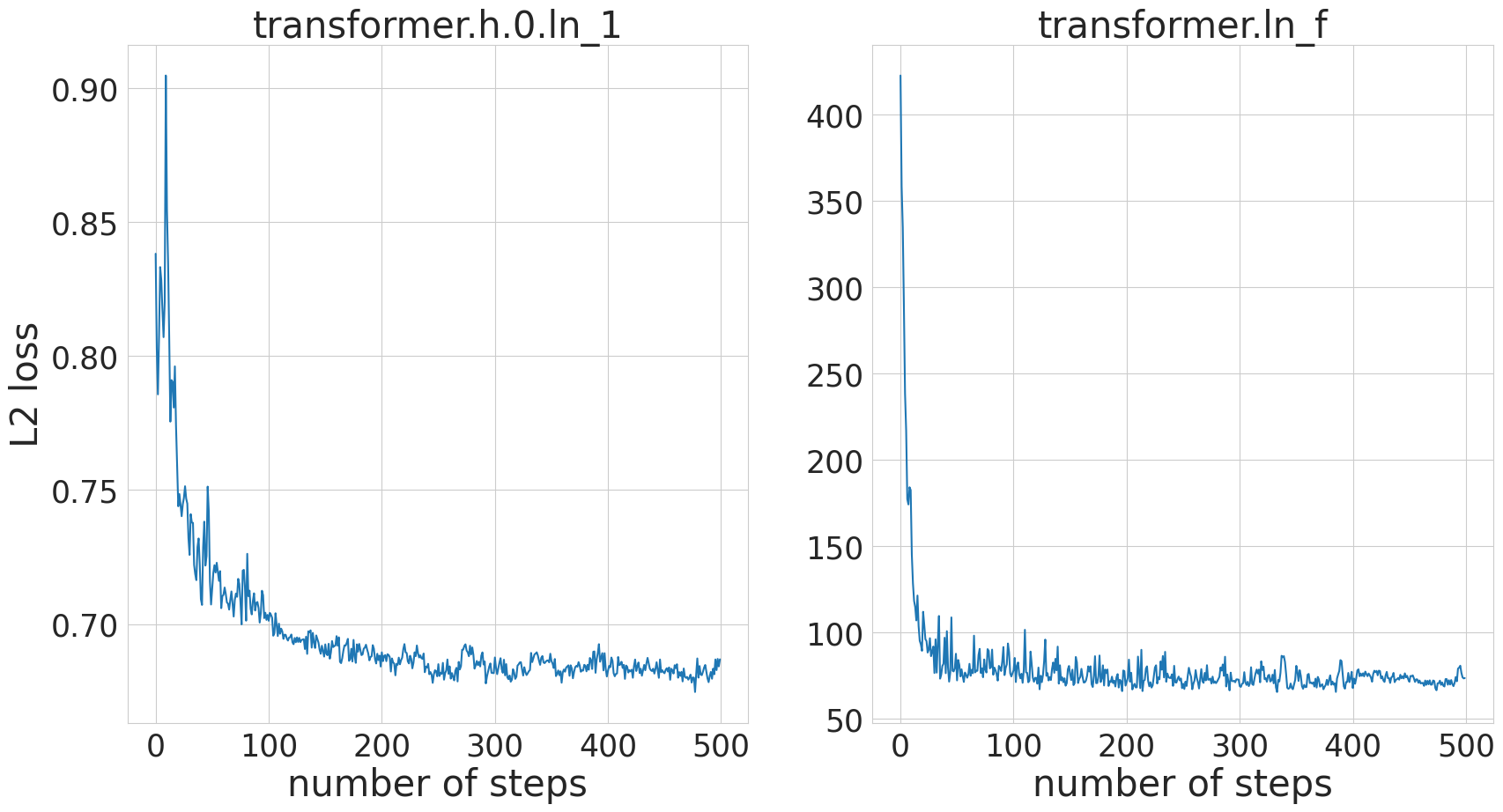}
    \caption{Block-wise calibration learning curves of the two selected Quadapter blocks in GPT-2.}
    \label{fig:learning_curve }
\end{figure}

\noindent{\textbf{Block-wise Calibration}}
We use 10 calibration data with the max time length (block size) of 512, 
yielding 5120 text tokens in total. The same calibration data is used for all the PTQ and QAT experiments.
The initial learning rate is 0.1, and it decays at the rate of 0.2 every 100 steps.
The training takes place for 500 steps with the Adam optimizer. 
Figure~\ref{fig:learning_curve } shows the convergence of training in two of the GPT-2 Quadapter blocks: the very first layer norm and the final layer norm. The total block-wise calibration phase takes approximately 2 minutes with RTX 2080 Ti, when training all the Quadatper blocks in GPT-2 sequentially from bottom to top.

\noindent{\textbf{End-to-end Fine-tuning}}
We use the initial learning rates 1e-5, 1e-3, and 1e-3 for
model parameters, Quadapter $\ba$, and quantization parameters $\btheta$, 
respectively. 
The learning rate linearly decreases to 0 over 10,000 training steps. 
The batch size is set to 4 with 
the max time length of 512.
All the QAT methods follow this training scheme.
After the completion of training, PPL is measured with the max time length of 1024. All of the PPL metrics reported in this work share this configuration.


\section{Implementation Details}
\noindent{\textbf{Quantization Implementation}}
The quantizer function $\tQ_\btheta$ is defined as follows: 
\begin{align}
    \tQ_\btheta(x) = s \cdot (clip(\nint*{\frac{x}{s} + o} , 0, 2^{b}-1) - o),  \\
    s = \frac{\theta_{max} - \theta_{min}}{2^b-1}, \quad o = \nint*{-\frac{\theta_{min}}{s}},
\label{eq:q_eq}
\end{align}
where $s$ and $o$ are the scale and offset, and $b$ is the target bit depth (8-bit in our case).
$\nint{\cdot}$ is a rounding function, and $clip(\bx, n, p)$ clamps all the values between $n$ and $p$ from the input $\bx$.

In the first calibration phase, $\theta_{min}$ and $\theta_{max}$ are obtained from the batch statistics at each inference step (i.e. dynamic quantization). 
In the second fine-tuning phase, the parameters are initially set based on the calibration data (i.e. static quantization), 
and trained afterwards.
When $\btheta$ is trained, the gradients are passed through the rounding function via straight-through-estimation.

\noindent{\textbf{Quadapter Implementation}}
The details of the actual application of Quadapter to GPT-2 is slightly different from the general form of Quadapter block in Equation~\ref{eq:main_1eq}. 
In the transformer block of GPT-2, $\bW_1$ denotes only the affine transformation
, but not the preceding normalization of the layer norm operation.
For example, we can define the layer norm operation as follows:
\begin{align}
    \by_{l} = \frac{\bx_{l} - \mu(\bx_{l})}{\sigma(\bx_{l})} 
    \odot\gamma + \beta,
\label{eq:layernorm}
\end{align}
then $\bW_1=\gamma, \bb_1=\beta$, and the input of Quadapter block is
$(\bx_l-\mu(\bx_l))/\sigma(\bx_l)$. The fused weight is computed as
$\bW_1'=\ba \odot \gamma$ with element-wise multiplication $\odot$.


\section{Constraint of Two Linear Layers}
While stating that Quadapter is applicable only to two consecutive layers of linear nature in Section 5, the main text omits the discussion of piece-wise linear activation function for brevity. For an operation $f$ that meets the following scaling-invariant condition:
\begin{align}
    f(sx) = sf(x),
\label{eq:piece-wise}
\end{align}
the identity relation between the scaling and the inverse-scaling of Quadapter still holds.
Therefore, it is possible to install Quadapter through piece-wise linear functions (e.g. ReLU, leaky ReLU, PReLU, etc.) as well.

Our future goal is to further expand the Quadapter applicability. We thus plan to investigate if Quadapter would be applicable even with an intervening nonlinear activation function (e.g. GeLU, tanh, etc.) by enhancing expressivity (i.e. setting up additional learnable scalar variables for the inverse scaling). In addition, by scaling all the tensors involved in a residual connection, we expect to apply Quadapter even in the presence of a residual connection in between the two target layers. 


\end{document}

%% file: macros.tex
\newcommand{\argmin}{\operatornamewithlimits{\arg \min}}

\newcommand{\bA}{{\mathbf{A}}}

\newcommand{\bx}{{\mathbf{x}}}

\newcommand{\bW}{\mathbf{W}}
\newcommand{\ba}{\mathbf{\alpha}}
\newcommand{\bb}{\mathbf{b}}
\newcommand{\tQ}{\textit{Q}}
\newcommand{\by}{\mathbf{y}}

\newcommand{\btheta}{{\boldsymbol \theta}}

\newcommand{\bbR}{\mathbb{R}}

\definecolor{orange}{RGB}{255,127,0}